\pdfoutput=1
\documentclass[11pt]{article}

\usepackage[preprint]{acl}

\usepackage{graphicx} % DO NOT CHANGE THIS
\usepackage{natbib}  % DO NOT CHANGE THIS AND DO NOT ADD ANY OPTIONS TO IT
\usepackage{caption} % DO NOT CHANGE THIS AND DO NOT ADD ANY OPTIONS TO IT
\usepackage{amsmath}
\usepackage{times}  % DO NOT CHANGE THIS
\usepackage{helvet}  % DO NOT CHANGE THIS
\usepackage{courier}  % DO NOT CHANGE THIS
\usepackage{graphicx} % DO NOT CHANGE THIS
\usepackage{booktabs}
\usepackage{amssymb}
\usepackage{CJKutf8}
\usepackage{tabularx}
\usepackage{array}
\usepackage{algorithm}
\usepackage{algorithmic}
\usepackage{amsmath}
\usepackage{newfloat}
\usepackage{listings}
\title{When Is a Steerable Concept Representation Real?\\Measurement Confounds in a Cross-Family Audit of Neuroscience Parallels in LLMs}

\author{
Yuqi Wu$^{1}$,  
Shengming Zhao$^{1}$,
\textbf{Jie Chen}$^{1}$\thanks{Correspondence, jc65\@ualberta.ca} \\
$^{1}$College of Biomedical Engineering, Fudan University
}

\begin{document}
\maketitle
\begin{abstract}
Large language models (LLMs) are increasingly reported to exhibit human-like neural and cognitive signatures, including concept cells, mental number lines, and cognitive maps. These claims often rely on linear probing and activation steering applied to a single model, yet both methods are highly sensitive to measurement choices. A reported parallel may therefore reflect the model, the measurement procedure, or both. We audit four representative neuroscience-inspired paradigms across 17 models from five families, spanning $0.6$B to $72$B parameters. Our main experiment examines the causal steerability of concept directions. With raw activation units and a fixed layer and coefficient, steerability appears to increase with model scale, resembling an emergent capability. However, this pattern is produced by an uncalibrated pipeline rather than by a claim established in the steering literature. The trend depends jointly on raw units, the readout metric, and the operating point; correcting any one of these removes it. With residual-norm-comparable interventions and held-out operating-point selection, concept steering remains significant at every scale, but shows no significant trend across the Qwen3 series, although the confidence interval does not rule out a moderate positive slope. The remaining results are mixed. A linear geographic world map is consistently decodable in every tested checkpoint up to $72$B. Number magnitude is strongly encoded, but whether individual neurons appear bell-shaped or monotonic depends on the selection criterion. Language-specific structure is localizable, but the direction of the cross-lingual asymmetry reverses under a different attribution method. These results suggest that the main constraint on AI neuroscience is not a lack of phenomena, but a lack of comparable measurements and adequate controls. We release the protocol, stimuli, and code.
\end{abstract}

\section{Introduction}

\begin{figure*}[t]
\centering
\includegraphics[width=0.88\textwidth]{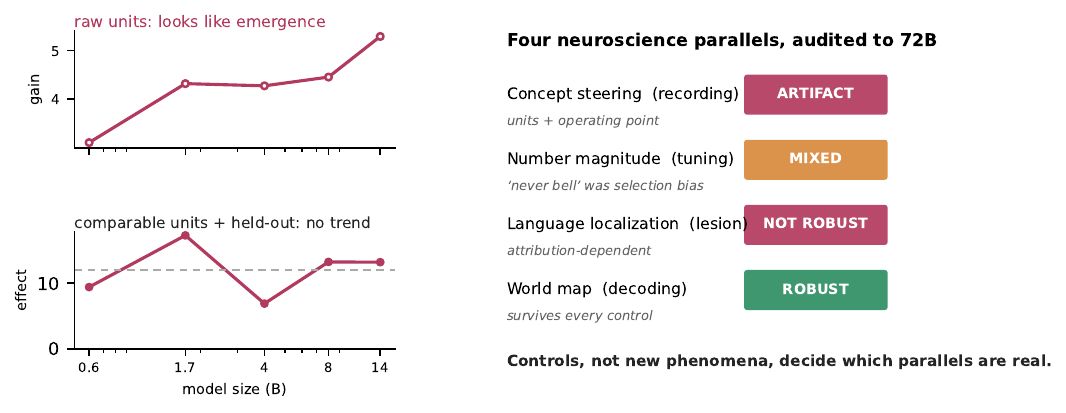}
\caption{\textbf{Same data, different ruler, different conclusion.}
\emph{Left:} a standard pipeline makes concept steerability appear to increase with model size in raw activation units. Once the intervention is normalized across models and the operating point is selected on held-out concepts, no significant scaling trend remains.
\emph{Right:} the four neuroscience-inspired phenomena examined in this study and the conclusions supported after control analyses.}
\label{fig:graphical}
\end{figure*}

A growing body of ``AI neuroscience'' treats large language models (LLMs) as systems whose internal representations can be recorded and perturbed. This literature has reported several parallels with biological cognition, including neurons selective for individual concepts \citep{quiroga2005invariant,templeton2024scaling}, linear representations of space and time \citep{gurnee2024language}, log-compressed number representations with Weber--Fechner behaviour \citep{numberrep2025,nieder2016neuronal}, and emotion or concept directions that causally influence model outputs \citep{sofroniew2026emotion,rimsky2024steering}. Together, these findings suggest that LLMs may exhibit representational structures resembling those observed in biological systems.

Many such results rely on two methods: \emph{linear probing}, which tests whether a variable can be decoded from activations, and \emph{activation steering}, which tests whether adding a direction changes model behaviour. Both are sensitive to analytical choices, including how units are selected, how intervention strength is defined, which layer is used, and whether specificity and null controls are included. Moreover, most studies examine a single model or a narrow range of model sizes. This makes it difficult to distinguish model-specific findings from broader regularities, or genuine effects from measurement artifacts. Similar concerns have already been raised for emergent capabilities, some of which disappear under alternative metrics \citep{schaeffer2023emergent}. Here, we examine the same problem in a causal representational setting.

We examine concept steering in depth and audit three additional neuroscience-inspired phenomena across 17 models from five families, spanning $0.6$B to $72$B parameters. Rather than asking whether these phenomena are ``real'' in an absolute sense, we test whether the associated claims remain stable under stronger controls and alternative measurement choices. For concept steering (\S\ref{sec:main}), the apparent increase with model scale is not measurement-invariant: it depends jointly on raw intervention units, the readout metric, and a fixed operating point, and disappears when any one of these is corrected. With residual-norm-comparable interventions and held-out operating-point selection, steering remains significant at every tested scale, but shows no significant trend across the Qwen3 series. We formalize these controls in an audit protocol (\S\ref{sec:methods}) combining held-out selection, residual-norm normalization, specificity and null controls, and cross-family replication. Applied across all four analyses, the protocol yields a cross-model taxonomy (\S\ref{sec:results}): linear geographic structure is consistently decodable up to $72$B; number magnitude is robustly encoded, but bell-shaped versus monotonic tuning depends on neuron selection; and language-specific structure is localizable, although the direction of the cross-lingual asymmetry changes with the attribution method. We further derive an efficient-coding account of monotonic magnitude representations under an idealized scalar model (App.~\ref{app:deriv}) and release the protocol, stimuli, and code.\footnote{Anonymized:
\url{https://anonymous.4open.science/r/ai-neurosci-audition-45CF} (de-anonymized upon acceptance).}

\section{Related Work}
\paragraph{Cognitive parallels in LLMs.} LLMs linearly encode space and time
\citep{gurnee2024language}, task-specific world models \citep{li2023othello}, truth
\citep{marks2024geometry}, and number magnitude, reproducing distance/size/ratio effects and a
log-compressed number line \citep{numberrep2025,dehaene2003neural,nieder2016neuronal}; emotion,
concept, and persona directions have been characterized and steered
\citep{sofroniew2026emotion,templeton2024scaling}. These works establish that the relevant
\emph{geometry} exists. We ask which \emph{causal} claims survive controls across families and scales.
\paragraph{Probing and steering.} Linear probes read variables off activations
\citep{alain2016understanding}, but a decodable variable need not be one the model uses, which is why
control-task baselines are needed \citep{hewitt2019control}. Steering methods add a direction to
change behaviour: difference-in-means and contrastive activation addition \citep{rimsky2024steering},
activation addition \citep{turner2023activation}, inference-time intervention \citep{li2023inference},
and representation engineering \citep{zou2023representation}; the linear representation hypothesis
\citep{park2024linear} underpins most causal claims, and factual edits localize to MLP weights
\citep{meng2022locating}. Crucially, the steering literature does \emph{not} claim that steerability
\emph{emerges} with scale: \citet{rimsky2024steering} report mixed effects at $13$B, and the one
systematic cross-scale study \citep{ali2025scaling} finds that CAA effectiveness \emph{diminishes}
with model size. The apparent emergence our case study examines is therefore a
natural default, not an established claim; because an unnormalized recipe can manufacture either an
apparent rise or an apparent decline, raw units are not a comparable measure of steerability across
scale. Superposition \citep{elhage2022superposition}, dictionary learning
\citep{bricken2023monosemanticity}, and cross-model convergence \citep{huh2024platonic} warn that
decodable directions need not be the units of computation, a warning we operationalize. The
reliability of steering vectors is itself an active concern, and our units argument gives one concrete
failure mode.
\paragraph{Neuroscience anchors.} Concept cells \citep{quiroga2005invariant}; place and grid
cells \citep{o1971hippocampus,hafting2005microstructure,constantinescu2016organizing};
the geometry of abstraction \citep{bernardi2020geometry}; efficient coding
\citep{laughlin1981simple}; and bell-shaped number neurons provide the biological referents
against which we score the models.

\section{Methods}\label{sec:methods}

\subsection{Models and grid}
We evaluate $17$ open-weight checkpoints spanning five developers,
from $0.6$ to $72$B parameters: Qwen3 at $\{0.6,1.7,4,8,14,32\}$B and Qwen2.5-$72$B; Llama-3.2 at
$\{1,3\}$B, Llama-3.1-$8$B, and Llama-3.1-$70$B; Phi-3.5-mini ($3.8$B) and Phi-4 ($14$B); Ministral-$8$B and
Mistral-Small-$24$B; and Gemma-2 at $\{9,27\}$B. Qwen3/Qwen2.5 and Phi-4 are base (pre-trained) checkpoints; Llama, Phi-3.5-mini,
Ministral, and Mistral-Small are instruction-tuned; Gemma-2 uses the \texttt{-it} variant. This
heterogeneity is deliberate: we test whether results depend on the fine-tuning stage.
Checkpoints up to $14$B run in bf16 on a single GPU. The $24$--$72$B models
run in $4$- or $8$-bit, and a bf16-versus-$8$-bit control at $32$B confirms that quantization does
not affect our conclusions (App.~\ref{app:details}). All interventions use forward hooks on the
residual stream and MLP layers.

\subsection{Recording and linear probes}
For population decoding we fit ridge probes with cross-validated regularization on the
residual-stream activation at a token of interest, and predict a target variable (for the world
map, the latitude and longitude of a city named in the prompt); probes are always scored on
held-out items. For single-unit analyses we record MLP neuron activations (the input to
\texttt{down\_proj}) and characterize each neuron's tuning to a stimulus variable.

\subsection{Activation steering and residual-norm normalization}
To steer a concept (for example, \emph{Paris}) we build a direction $d$ as the mean residual of
concept-bearing sentences minus a neutral baseline at a chosen layer, and inject $\alpha d$ into
the residual stream. We read the effect on neutral prompts as the change in a target-versus-control
log-probability contrast (for example, Paris-tokens versus Tokyo-tokens). Steering coefficients
are conventionally reported in \emph{raw} activation units and are not normalized across models
\citep{rimsky2024steering}. Because the residual-stream norm $\|h\|_\ell$ grows with model size,
we express strength as a fraction of that norm, $\alpha = c\,\|h\|_\ell$, which makes the
intervention comparable across scales. Concept and control lists, stimuli, and probes are given in
App.~\ref{app:stim}.

\subsection{Specificity, directional, and null controls}
We standardize four choices that, left implicit, invalidate cross-model comparison. \textbf{(R1)
Held-out selection and evaluation}: any neuron, direction, or hyperparameter is \emph{selected} on
one split and \emph{scored} on a disjoint one. \textbf{(R2) Comparable intervention units}: steering
directions are \emph{unit-normalized} ($\hat d = d/\|d\|$) and their strength is the residual-norm
fraction $\alpha = c\,\|h\|_\ell$, so the actual injection is
$h' = h + c\,\|h\|_\ell\,\hat d$
and its norm is exactly $c\,\|h\|_\ell$, a fixed fraction of the residual regardless of $\|d\|$.
The naive (uncorrected) pipeline omits both normalizations: it injects the raw mean-difference
direction at a raw scalar coefficient, so the injection norm depends unpredictably on both $\|d\|$ and
the model. \textbf{(R3) Specificity and directional controls}: a causal claim for a direction $d$
(target $T$ versus control $C$) must show that $d$ raises $\log p(T)-\log p(C)$, that the control
direction $d_C$ reverses it, and that random directions do not. \textbf{(R4) Nulls and confidence
intervals}: scores are compared against shuffled or pseudo-concept nulls and reported with bootstrap
confidence intervals. A concept \emph{passes} when all three R3 tests hold against the R4 null, and the
\emph{pass-rate} is the fraction of concepts that pass; alongside it we report the mean continuous
specificity effect $\log p(T)-\log p(C)$, which we treat as the primary metric (the pass-rate is
grid-sensitive).

\subsection{Held-out operating-point selection}
The steering effect depends jointly on the intervention layer and the strength $c$. Rather than fix
these by hand, we select them the way the world-map probe selects its layer and regularizer: by
held-out performance. We scan a (layer, strength) grid on part of the concepts and score the
disjoint rest. For the main Qwen3 analysis we use $24$ concepts across five categories (cities,
persons, objects, animals, abstractions), a fine grid, and a four-fold split; the $24$--$72$B models
use a coarser grid (App.~\ref{app:sens}).

\subsection{Lesioning and attribution}
To test language localization we locate language-selective neurons, ablate the top fraction, and
measure the change in held-out next-token loss on each language. Neurons are ranked by
gradient$\times$activation attribution; as a robustness check we re-rank them by activation-magnitude
difference between languages (App.~\ref{app:sens}). On models of $14$B and above the attribution
backward pass exceeded memory because it allocated a full weight-gradient buffer; freezing all
weights and marking only the recorded activations \texttt{requires\_grad} removes that buffer while
leaving the attribution numerically identical, so all $17$ models are covered.

\subsection{Statistical procedures}
Effects carry bootstrap $95\%$ confidence intervals, resampled over concepts for steering and over sentences for lesioning. For the steering scaling question we regress the per-concept held-out effect on $\log_2$ model size and report the slope with a bootstrap CI (a trend test). We restrict this regression to the Qwen3 ladder to hold family fixed; even so, its checkpoints are separate training runs rather than one model rescaled, so the slope estimates a trend across an empirical sequence of checkpoints, not a controlled effect of parameter count, and inherits whatever data- and recipe-differences distinguish those runs. For magnitude
shape we correlate each neuron's digit and spelled-out-word tuning curves (a cross-format test).
Null distributions match the structure of each test (shuffled labels, pseudo-concepts).

\section{Experiments and Results}\label{sec:results}

\subsection{The ``concept steerability emerges with scale'' finding is a measurement artifact}\label{sec:main}

\begin{figure*}[t]\centering
\includegraphics[width=0.75\textwidth]{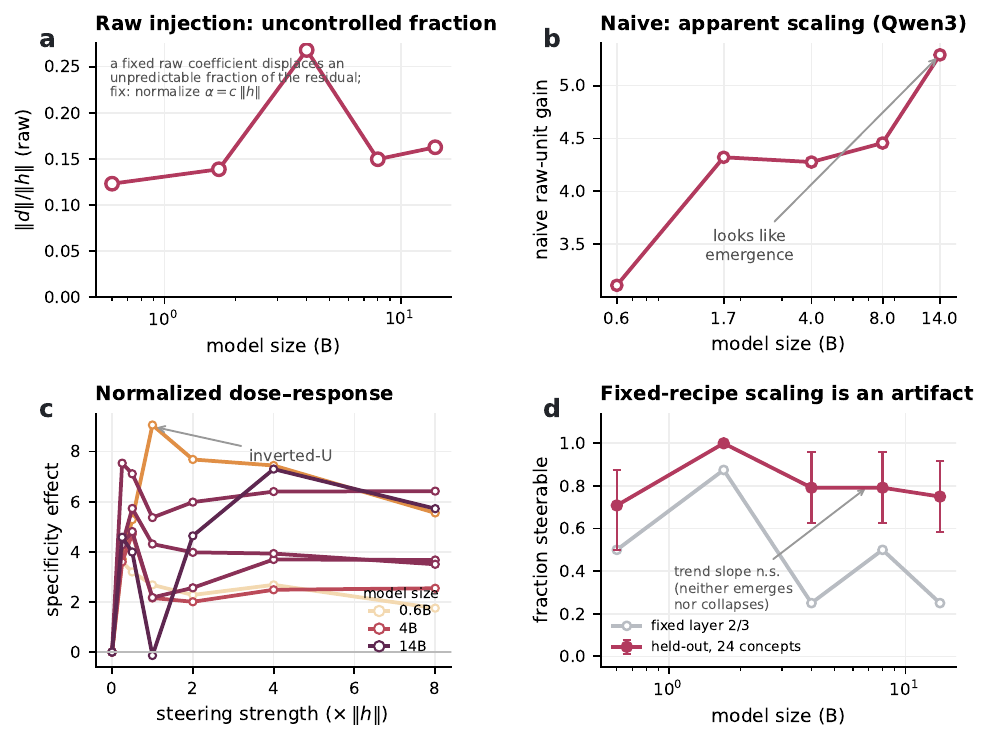}
\caption{\textbf{Main experiment: the ``concept steerability emerges with scale'' effect is a
measurement artifact.} (a) The units confound: the raw injection fraction $\|d\|/\|h\|$ varies
non-monotonically across the Qwen3 ladder, so a fixed raw coefficient is an uncontrolled
intervention. (b) In raw units the effect nonetheless appears to grow monotonically with scale, an
apparent emergent capability. (c) Under residual-norm normalization with specificity control, the
strength dose-response is an inverted-U, not monotone growth (colour $=$ size). (d) At a fixed,
hand-picked layer ($2/3$; grey) the pass-rate is erratic, but selecting the operating point on
held-out concepts (red; $24$ concepts, bootstrap $95\%$ CI) leaves steering significant at every
scale with no significant scaling trend. Red and grey come from different selection pipelines and are
not point-for-point comparable.}
\label{fig:main}
\end{figure*}

\paragraph{An apparent scaling law.} Steering coefficients are conventionally reported in raw,
unnormalized units \citep{rimsky2024steering}. Under this default pipeline, a raw mean-difference
direction, a raw steering coefficient, and a first-token target probability, the effect grows
\emph{monotonically} across the Qwen3 ladder (Fig.~\ref{fig:main}b): the gain rises from $0.6$B to $14$B,
with exactly the shape one would report as an emergent capability.

\paragraph{Why it is an artifact.} A raw steering coefficient is not a comparable unit of
intervention. Across the Qwen3 ladder the residual-stream norm $\|h\|_\ell$ and the raw steering
direction $\|d\|$ both vary substantially and non-monotonically with model size, and the fraction of
the residual that a raw injection displaces, $\|d\|/\|h\|$, swings between $0.12$ and $0.27$ with no
clean dependence on scale (Fig.~\ref{fig:main}a). A fixed raw coefficient therefore delivers a different,
unpredictable intervention in each model, and any scaling law read from raw units reflects this
incomparability rather than steerability. Normalizing the injection to the residual norm
($\alpha=c\|h\|$) restores comparability. The argument is scale-independent: it shows the ruler is
miscalibrated, not that larger models are less steerable, and the same confound distorts a $72$B
comparison identically.

\paragraph{Dose-response and layer sensitivity.} Under R2 the specificity-controlled effect is not
monotone in strength but follows an inverted-U (Fig.~\ref{fig:main}c): it rises, peaks at
low-to-intermediate $c$, then declines as stronger steering degrades the output distribution. The
effect is also sharply layer-dependent (Fig.~\ref{fig:layer}), rising with depth and peaking in late
layers, so a single fixed layer, as used in most steering papers, can under- or over-state the
effect. Any single-number ``steering effect'' is thus a point on two hidden axes.

\paragraph{A second confound: the operating point.} With the specificity, directional, and null
controls in place (R3--R4: the target direction must raise $T$ over control $C$, the control
direction must reverse it, random directions must not, all against a bootstrap null), a genuine
direction-specific effect is present at every scale. Its apparent \emph{scaling}, however, is set by
where the intervention is applied. At a fixed, hand-picked operating point (layer $2/3$, strength
$1{\cdot}\|h\|$) the specificity pass-rate is erratic across the Qwen3 ladder
($0.50, 0.88, 0.25, 0.50, 0.25$; Fig.~\ref{fig:main}d, grey), which naively reads as a collapse. Selecting
the operating point on held-out concepts (Methods) and expanding to $24$ concepts across five
categories with bootstrap CIs gives a stable picture: the held-out pass-rate is
$0.71, 1.00, 0.79, 0.79, 0.75$ over $0.6$--$14$B, each $95\%$ CI above chance; the continuous
specificity effect is significantly positive at every scale; and the slope of effect against
$\log_2$ size is $+0.31$ per doubling, $95\%$ CI $[-0.11,+0.73]$---no significant scaling trend,
though the CI does not exclude a moderate positive slope
(Fig.~\ref{fig:main}d, red). We read this as the absence of a detectable trend across this particular sequence of checkpoints, not as evidence that steerability is scale-invariant: the Qwen3 sizes are independently trained runs, so a homogeneous size axis is itself an assumption, and the test remains underpowered against the moderate slopes the interval still admits. Every model selects layer $\approx\!0.8$, never the hand-picked $2/3$, and
concrete categories (objects, persons, cities) steer more strongly than abstract ones. No fixed
operating point is comparable across models: both the apparent emergence and the apparent collapse
are artifacts of that choice, removed only by selecting the operating point on held-out data.

\paragraph{Factor decomposition.} To isolate which measurement choice drives the apparent scaling, we
hold the metric (specificity contrast) and concept set ($24$ concepts) constant and vary two factors
independently: unit normalization (raw vs.\ $\|h\|$-normalized) and operating-point selection (fixed
$L{\cdot}2/3$ vs.\ held-out grid search). None of the four resulting cells shows a significant scaling
trend (Fig.~\ref{fig:factor}): raw+fixed slope $+0.22$, $95\%$ CI $[-0.02,+0.62]$; normalized+held-out
slope $+0.31$, CI $[-2.87,+4.05]$; the other two are similarly flat. (The wider CIs in
the decomposition relative to the main analysis reflect fewer probe prompts per concept---three
vs.\ six---reducing per-concept precision; both use the same five models.) The apparent scaling in the
naive pipeline (Fig.~\ref{fig:main}b) therefore rests not on a single confound but on the \emph{compound}
of several uncalibrated choices---units, readout metric, and operating point---any one of which, when
corrected, removes the trend.

\paragraph{The causal reading.} Two observations sharpen it. First, the most concept-selective
single MLP neurons are not \emph{sufficient} levers: clamping one degrades output rather than
inducing the concept, whereas a population direction does steer, an instance of the
decodable$\neq$used gap \citep{elhage2022superposition}; Fig.~\ref{fig:steerexample} (App.~\ref{app:qual})
shows real steered generations, in which a \emph{Paris} direction bends a neutral story toward the
Eiffel Tower while staying fluent. Second, held-out evaluation matters:
in-sample single-neuron concept selectivity ($d'\!\approx\!6$ on a representative model) roughly
halves out of sample ($d'\!\approx\!2$), with several concepts falling below the pseudo-concept
null.

\subsection{Steerability across scale and a quantization control}\label{sec:scale}
Extending the held-out analysis to the large models (Table~\ref{tab:results}), the held-out
\emph{effect} is significantly positive at every large scale, with a bootstrap $95\%$ CI above $0$
for all five (Gemma-2-27B $+22.9\,[+18.5,+27.1]$, Qwen2.5-72B $+7.1\,[+6.0,+8.3]$, Qwen3-32B $+4.9\,[+2.7,+7.2]$, Mistral-24B
$+5.5\,[+4.5,+6.4]$, Llama-70B $+4.5\,[+3.0,+6.1]$), so steering does not vanish at scale. We rest
this on the continuous effect and its CI rather than the pass-rate, which is grid-sensitive: the
coarse large-model grid yields lower absolute pass-rates ($0.38$ at $32$B) that are not comparable to
the fine $14$B scan. A bf16-versus-$8$-bit control at $32$B leaves the effect essentially unchanged
($+4.9$ versus $+4.3$, overlapping CIs), so quantization does not drive the result. Because that grid
is coarse and mixes families, we read only significance and precision-invariance at $24$--$72$B; the
``no detectable trend'' statement is reserved for the dense Qwen3 ladder to $14$B. Table~\ref{tab:results}
reports every per-model number behind the four experiments.

\begin{table*}[t]\centering\small
\setlength{\tabcolsep}{5pt}
\caption{\textbf{Per-model results across the $0.6$--$72$B grid, all four experiments.} Steering:
held-out specificity effect and pass-rate. Number: bell-neuron fraction (shape-agnostic selection)
and digit-vs-word cross-format tuning correlation over \emph{all} well-tuned bell neurons in the
model (not limited to the top $30$; $>\!0.5$ indicates partly format-invariant magnitude tuning). Lesion (grad$\times$activation): Chinese/English selectivity. Map: cross-validated geographic $R^2$
(latitude/longitude). \textbf{Bold} marks a statistically significant selectivity (bootstrap $95\%$
CI excluding $0$); we bold only the lesion columns, where significance varies model to model, since
the steering effect is significant at every scale and the number and map columns are point estimates.
Pass-rate is the fraction of concepts passing the specificity, directional, and null controls (R3--R4).
The five $\ge\!24$B rows use a coarse steering grid, so their absolute pass-rates are not comparable to
the fine $\le\!14$B scan. Every cell is filled (all $17$ models $\times$ all four experiments). The world
map is the one column consistent across all models; every other headline is scale-, family-,
or selection-dependent.}
\label{tab:results}
\begin{tabular}{lccccccccc}
\toprule
& & & \multicolumn{2}{c}{Steering} & \multicolumn{2}{c}{Number} & \multicolumn{2}{c}{Lesion (sel.)} & Map \\
\cmidrule(lr){4-5}\cmidrule(lr){6-7}\cmidrule(lr){8-9}\cmidrule(lr){10-10}
Model & B & Fam. & eff & pass & bell\% & x-fmt $r$ & ZH & EN & $R^2$ \\
\midrule
Qwen3-0.6B & 0.6 & Qwen & +9.4 & 0.71 & 10 & +0.55 & \textbf{+3.3} & \textbf{+0.7} & 0.53/0.67 \\
Llama-3.2-1B & 1 & Llama & +11.0 & 0.92 & 17 & +0.64 & \textbf{+6.0} & -1.0 & 0.60/0.61 \\
Qwen3-1.7B & 1.7 & Qwen & +17.2 & 1.00 & 3 & +0.45 & \textbf{+2.8} & +0.0 & 0.62/0.66 \\
Llama-3.2-3B & 3 & Llama & +13.1 & 0.96 & 20 & +0.64 & \textbf{+8.2} & -1.5 & 0.54/0.46 \\
Phi-3.5-mini & 3.8 & Phi & +29.4 & 0.83 & 13 & +0.68 & \textbf{+9.1} & +0.2 & 0.66/0.61 \\
Qwen3-4B & 4 & Qwen & +6.9 & 0.79 & 0 & +0.46 & \textbf{+3.2} & +0.2 & 0.52/0.58 \\
Llama-3.1-8B & 8 & Llama & +11.0 & 0.83 & 23 & +0.65 & \textbf{+9.8} & -0.7 & 0.46/0.64 \\
Ministral-8B & 8 & Mistral & +7.0 & 0.96 & 17 & +0.60 & \textbf{+6.8} & \textbf{+1.1} & 0.53/0.59 \\
Qwen3-8B & 8 & Qwen & +13.2 & 0.79 & 0 & +0.38 & \textbf{+1.9} & -0.5 & 0.50/0.49 \\
Gemma-2-9B & 9 & Gemma & +21.0 & 0.96 & 23 & +0.69 & \textbf{+4.6} & +0.2 & 0.60/0.47 \\
Qwen3-14B & 14 & Qwen & +13.1 & 0.75 & 0 & +0.01 & \textbf{+1.0} & +0.1 & 0.52/0.61 \\
phi-4 & 14 & Phi & +17.8 & 0.96 & 27 & +0.59 & \textbf{+4.6} & -0.7 & 0.61/0.61 \\
Mistral-Small-24B & 24 & Mistral & +5.5 & 0.92 & 40 & +0.70 & \textbf{+7.5} & -1.0 & 0.55/0.67 \\
Gemma-2-27B & 27 & Gemma & +22.9 & 0.54 & 0 & +0.51 & \textbf{+7.1} & \textbf{+0.3} & 0.59/0.68 \\
Qwen3-32B & 32 & Qwen & +4.9 & 0.38 & 0 & +0.36 & \textbf{+2.8} & \textbf{+3.7} & 0.53/0.43 \\
Llama-3.1-70B & 70 & Llama & +4.5 & 0.62 & 43 & +0.69 & \textbf{+8.8} & -4.2 & 0.56/0.67 \\
Qwen2.5-72B & 72 & Qwen & +7.1 & 0.88 & 0 & +0.64 & \textbf{+2.3} & \textbf{+0.5} & 0.58/0.45 \\
\bottomrule
\end{tabular}
\end{table*}

\subsection{Magnitude coding and number neurons}\label{sec:mag}
\begin{figure}[t]\centering
\includegraphics[width=\columnwidth]{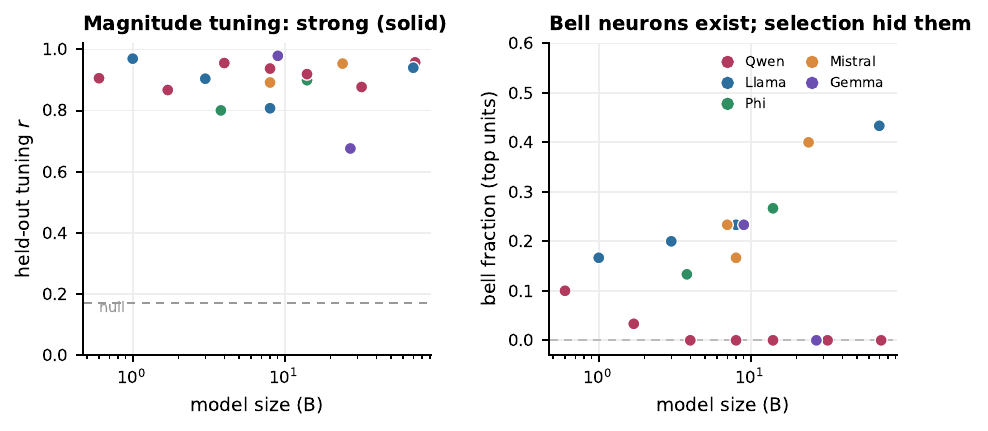}
\caption{\textbf{Magnitude: strong tuning, mixed shape.} (a) Held-out magnitude tuning is
strong in every model, to $72$B. (b) Under \emph{shape-agnostic} neuron selection, bell-shaped
number neurons appear, common in Llama/Phi/Mistral and rare in Qwen; the linear-correlation
selection that returned ``$0\%$ bell'' was biased.}
\label{fig:mag}
\end{figure}

Every
model contains strong magnitude tuning: the best number-tuned neuron reaches held-out
$r=0.68$--$0.98$, far above null, in all $17$ models up to $72$B (Fig.~\ref{fig:mag}a; the $32$B
bf16-versus-$8$-bit control is identical). Whether these neurons look bell-shaped or monotonic,
however, is entirely a matter of selection. Ranking neurons by \emph{linear} correlation with
number---the natural first choice, and the one that recovers biological number-neuron studies---
returns only \emph{monotonic} units, because a linear criterion cannot rank an interior-peaked tuning
curve highly. A \emph{shape-agnostic} criterion (a held-out quadratic fit; App.~\ref{app:sens})
tells a different story: bell-shaped (interior-peak, negative quadratic coefficient) number neurons
are present in every model (hundreds to thousands among all well-tuned units), and are common among
the top $30$ in Llama, Phi, and Mistral, mixed in Gemma, and rare-to-absent in the top $30$ for Qwen
(Fig.~\ref{fig:mag}b; fractions are threshold-sensitive, App.~\ref{app:sens}). Are these genuine
numerosity units or digit-token detectors? A cross-format test asks whether a neuron keeps its
tuning curve when the number is a digit (``$17$'') or a spelled-out word (``seventeen''). The bell
units are \emph{partly format-invariant} in Llama, Phi, and Mistral (cross-format tuning correlation
$0.59$--$0.70$, above each model's random-neuron baseline of $0.30$--$0.51$, with $73$--$83\%$ of bell
units keeping $r>0.5$; the margin over random is large in Phi but modest in Ministral;
App.~\ref{app:sens}), a split that holds to $72$B (Llama-70B and Mistral-24B $r\approx0.7$, Qwen
monotonic-dominant), which indicates numerical-magnitude tuning beyond pure token identity, though
not fully invariant and weaker in Qwen. Fig.~\ref{fig:numberexample} (App.~\ref{app:qual}) shows a real
bell-shaped neuron whose digit and word curves nearly coincide. Magnitude is therefore strongly encoded, but the code is
\emph{mixed}: monotonic units alongside partly format-invariant bell units, in a family-dependent
mix rather than a clean monotonic-only line. An efficient-coding argument
(App.~\ref{app:deriv}) shows why a monotonic code would be \emph{optimal} for frequency-weighted
discrimination \citep{laughlin1981simple,dehaene2003neural,numberrep2025}, but
that is a normative prediction the models only partly follow, not a description of what they do.

\subsection{Language localization}\label{sec:lesion}
\begin{figure}[t]\centering
\includegraphics[width=\columnwidth]{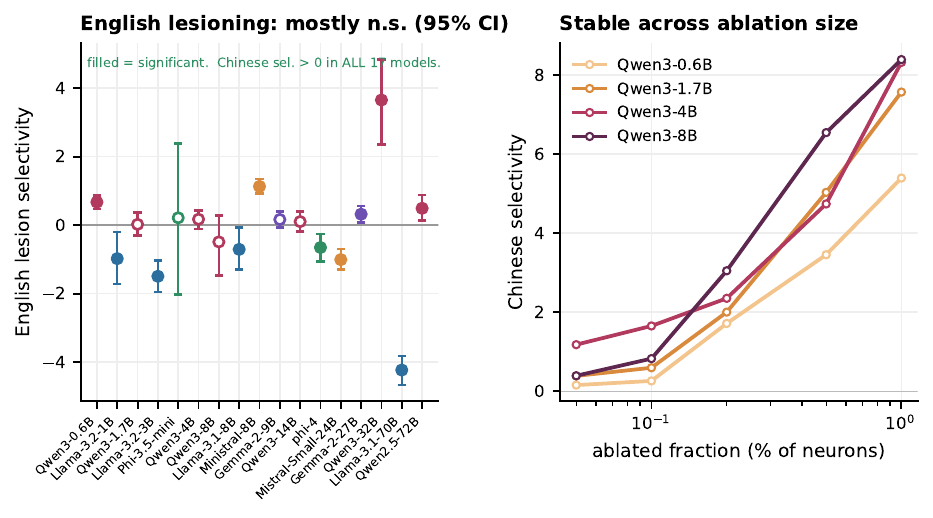}
\caption{\textbf{Language lesioning (gradient$\times$activation): only Chinese is reliably
localizable, and this asymmetry itself flips under a different attribution (App.~\ref{app:sens}).}
(a) English lesion selectivity with bootstrap $95\%$ CIs ($44$ sentences/language); filled $=$ CI
excludes $0$. It is null in $12/17$ models and significantly positive in only five, so there is no
reliable English dissociation. Chinese selectivity (not shown) is significantly $>0$ in \emph{all}
$17$. (b) The Chinese effect is stable across ablation fraction.}
\label{fig:lesion}
\end{figure}

Locating language-selective neurons by gradient$\times$activation attribution, ablating the top
fraction, and bootstrapping the held-out loss over $44$ sentences per language
(App.~\ref{app:sens}), the \emph{Chinese} half is robust: Chinese-selective ablation hurts Chinese
significantly more than English in \emph{all} $17$ tested models ($95\%$ CI $>0$ everywhere). The
\emph{English} half is not. English selectivity is statistically indistinguishable from zero in $12$
of $17$ models (CI straddles $0$; Fig.~\ref{fig:lesion}a) and significantly positive in only five
(Qwen3-0.6B, Ministral-8B, Qwen3-32B, Qwen2.5-72B, Gemma-2-27B). A full double dissociation, with both halves
significant and positive, therefore holds in only $5$ of $17$ models. However, the formal
ablated-language$\times$test-language interaction (zh\_sel $+$ en\_sel, with CIs combined in quadrature)
is significantly positive in all $17$ models: language-selective ablation hurts the targeted language
more than the non-targeted one everywhere, even when the per-language selectivities individually fail
to reach significance. Even this asymmetry is \emph{attribution-dependent}, and this is our cleanest single
instance of measurement-dependence. Gradient$\times$activation ranks a neuron by its \emph{causal}
importance (how much perturbing it changes the loss), whereas activation magnitude ranks it by how
strongly it \emph{fires} for a language; the two diverge because a unit can be highly active for a
language without being causally load-bearing, the same decodable$\neq$used gap that undercuts single
directions (\S\ref{sec:main}). Under the activation-magnitude ranking the asymmetry flips: Chinese is
significant in only $7/9$ models (two Qwen fail) while English becomes significant in $7/9$
(App.~\ref{app:sens}). The only claim robust to \emph{both} the bootstrap and a change of attribution
is the weakest one: \emph{some} language-specific structure is localizable in most models, but which
language, how cleanly, and in which direction depend on which notion of ``importance'' the analyst picks. At scale, gradient$\times$activation keeps Chinese significant to $72$B. English
selectivity is significant in both large Qwen models but in neither Mistral-24B nor Llama-70B, and
its magnitude is erratic (Qwen3-32B $+3.7$ vs.\ Qwen2.5-72B $+0.5$); we read this as further
model-dependence, not a scale effect. A lesion-fraction sweep (Fig.~\ref{fig:lesion}b) confirms that the
Chinese effect is stable across ablation size, and Fig.~\ref{fig:lesionexample} (App.~\ref{app:qual})
shows a single-sentence example: ablating Chinese neurons raises Chinese surprisal by $+10.7$ nats
while sparing English ($+0.5$).

\subsection{A consistent linear world map}\label{sec:map}
\begin{figure}[t]\centering
\includegraphics[width=0.72\columnwidth]{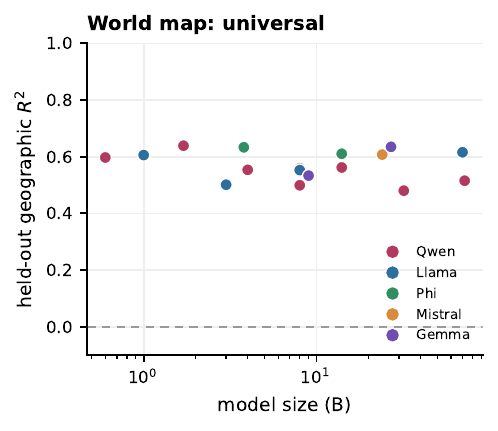}
\caption{\textbf{Consistent linear world map} across five families, $0.6$--$72$B (null $\approx 0$).}
\label{fig:map}
\end{figure}

Ridge probes (cross-validated regularization, held-out cities) recover latitude and longitude from
the city-token residual with $R^2\!\approx\!0.43$--$0.68$ in all $17$ models, near zero under label
shuffling (Fig.~\ref{fig:map}), replicating \citet{gurnee2024language} across five families and the
full $0.6$--$72$B range. This is the one parallel consistent across all tested checkpoints
(Table~\ref{tab:tax}); Fig.~\ref{fig:mapexample} (App.~\ref{app:qual}) plots the recovered coordinates,
which reproduce global geography. Its robustness is not a counterexample to our thesis but a
confirmation of it: the world map is the only one of the four tasks that is \emph{pure decoding},
with no intervention, no single-neuron selection, and no operating-point choice, so none of the three
artifact classes (unit incomparability, operating-point drift, selection bias) can reach it. Rigor is
decisive exactly where selection and intervention enter, and inert where they do not.

\begin{table}[t]\centering\small
\caption{Audit taxonomy ($17$ models, $0.6$--$72$B). Each shape, scaling, and family claim depends on
a measurement choice; only the linear world map is consistent across all tested checkpoints.}
\begin{tabular}{@{}p{2.9cm}p{1.2cm}p{2.6cm}@{}}\toprule
Parallel (method) & Verdict & What a control revealed\\\midrule
Concept steering ``emerges with scale'' (recording + intervention) & \textbf{artifact} & units + arbitrary operating point\\
Magnitude ``monotonic code'' (tuning) & \textbf{partial} & ``never bell'' was selection bias\\
World map (decoding) & \textbf{consistent} & ---\\
Language double dissociation (lesion) & \textbf{not robust} & attribution- \& sample-dependent\\
\bottomrule\end{tabular}\label{tab:tax}
\end{table}

\section{Discussion}\label{sec:disc}
Across four experiments (Table~\ref{tab:tax}) a single pattern holds: rigor is \emph{decisive}
exactly where selection and intervention enter (concept steering, lesions) and \emph{inert} for
purely correlational decoding (the world map), which is consistent to $72$B. Each headline parallel
weakens under its matching control. The steering ``emergence'' rests on two confounds, raw units and
an arbitrary operating point; choosing that point on held-out data leaves steerability significant at
every scale with no detectable trend. The magnitude ``monotonic code'' is a selection effect:
shape-agnostic selection finds bell-shaped neurons in every model. The lesion ``double dissociation''
survives neither a bootstrap nor a change of attribution method; the formal interaction is
significant everywhere, but whether the per-language halves individually hold depends on method
and model. This is not an argument that parallels between LLMs
and brains are false: geography is robustly mapped, magnitude robustly encoded, concepts genuinely
steerable, and Chinese reliably localizable. It is an argument that the field's binding constraint is
\emph{comparability and controls}: an unaudited probe or intervention tends to over-claim, and a
single arbitrary knob---a layer, a selection rule, a sample size---can fabricate a scaling law, a
coding shape, or a functional map.

\paragraph{Limitations.}
Although all experiments extend to $72$B, the densest scale analysis is limited to the Qwen lineage up to $14$B; larger models use coarser grids and quantized inference, though a $32$B bf16-versus-$8$-bit control shows closely matched results. The $24$-concept steering set limits precision, and broader coverage would better constrain scaling trends. Number-neuron results remain partly format-dependent, and non-symbolic stimuli would better separate magnitude from symbol identity. Family and scale effects are also not fully separable, since model families do not share a common size grid. Accordingly, the absence of a significant steering trend should be interpreted as no detectable trend rather than evidence of no trend. Construct validity could be strengthened through leave-name-out steering and non-lexical evaluation, while mean ablation and larger language test sets would further validate the lesion results. Finally, residual-norm normalization improves comparability but does not guarantee equal functional dose across models, and our analysis is limited to representational rather than dynamical phenomena.

\section{Conclusion}
Treating LLMs as objects of neuroscience is productive only with neuroscience-grade controls.
A deep case study shows an apparent causal scaling law to be an artifact of intervention units and an
arbitrary operating point. A cross-generation audit to $72$B then finds that, of three further
parallels, only the linear world map is consistent across all tested checkpoints: the magnitude ``monotonic code''
was a selection bias, and the language ``double dissociation'' does not survive bootstrap and
attribution-method controls. We release the protocol, grid, stimuli, and code so that future ``LLMs
also show $X$'' claims can be scored, not merely asserted.

\bibliography{main}
\appendix
\section{Derivation: the magnitude code is the CDF of number frequency}\label{app:deriv}
Let a model represent number $n$ by a scalar coordinate $\phi(n)\in[0,1]$ read out with fixed
resolution (additive noise $\sigma$ in $\phi$-space); numbers occur with frequency $p(n)$ and
training rewards discriminating them in proportion to $p$. The number of distinguishable
levels near $n$ is $\propto \phi'(n)/\sigma$, so expected discriminability
$\int p(n)\,\log\!\big(\phi'(n)/\sigma\big)\,dn$ is maximized subject to $\int\phi'(n)\,dn=1$
by the Lagrange condition $p(n)/\phi'(n)=\text{const}$:
\[
\phi'(n)\propto p(n)\qquad\Longrightarrow\qquad \phi(n)=\!\!\sum_{m\le n}\!p(m)=F(n),
\]
the CDF of $p$ (histogram equalization / optimal companding; \citealp{laughlin1981simple}).
For heavy-tailed text statistics $p(n)\!\propto\!1/n$, $\phi(n)\!\propto\!\log n$: a monotone
log-compressed line with constant Weber fraction $\Delta n/n$. Within this idealized model
(scalar code, additive read-out noise, reward $\propto$ frequency) the optimum is monotone for
any $p>0$, so a bell-shaped (non-monotone) code is sub-optimal. This is a statement about the
idealized population \emph{coordinate}, not about single-neuron tuning: our shape-agnostic
analysis (\S\ref{sec:mag}) shows individual neurons \emph{can} be bell-shaped, so the
derivation is a normative prediction of the optimal code, not a description of the LLM's units.

\section{Stimuli, concept lists, and controls}\label{app:stim}
\textbf{Concept steering.} The initial specificity analysis uses $8$ concepts in $4$ matched pairs:
\emph{Paris/Tokyo}, \emph{Einstein/Shakespeare}, \emph{Moon/Ocean}, \emph{Piano/Dinosaur}.
The held-out scaling analysis (\S\ref{sec:main}) expands to $24$ concepts in $12$ matched pairs across
five categories (cities, persons, objects, animals, abstract: e.g.\ Paris/Tokyo, Einstein/Shakespeare,
Piano/Dinosaur, Eagle/Dolphin, Freedom/Justice). Each concept's steering direction is built from
$\sim$3--8 sentences in Chinese and English, mixing named and description-only mentions (e.g.\ ``the
city where the Eiffel Tower stands''); the neutral baseline is $10$ topically neutral sentences.
Target/control word sets per concept (e.g.\ Paris: \{Paris, France, Eiffel, French\}) are scored by
full-sequence log-probability on $6$ neutral read-out prompts (``My favourite\ldots is'', etc.).
Bootstrap CIs are over concepts (maintaining concept identity across scales); random-direction
baselines use $5$ directions.
\textbf{Magnitude.} ``The number is $n$'', $n\!=\!1$--$40$; digit-token activation; best neuron
by train-split $|r|$, scored on the held-out split, $5$ random splits; monotonic vs.\ bell by
comparing linear and quadratic fit $R^2$ (bell if quadratic gain $>0.15$).
\textbf{Map.} $45$ world cities with approximate integer coordinates; ``The city of \{name\}'',
last-token residual; ridge with $\alpha$ chosen by inner CV; $5$-fold held-out $R^2$;
label-shuffle null ($200$ shuffles); RSA against great-circle distance.
\textbf{Lesion.} Parallel Chinese/English sentence sets (train/test split); attribution
$=$ mean $|a\cdot\partial_a\mathrm{NLL}|$ per neuron; ablate the top fraction to zero;
held-out loss change vs.\ equal-size random ablation.

\section{Sensitivity and robustness}\label{app:sens}
\begin{figure*}[t]\centering
\includegraphics[width=0.88\textwidth]{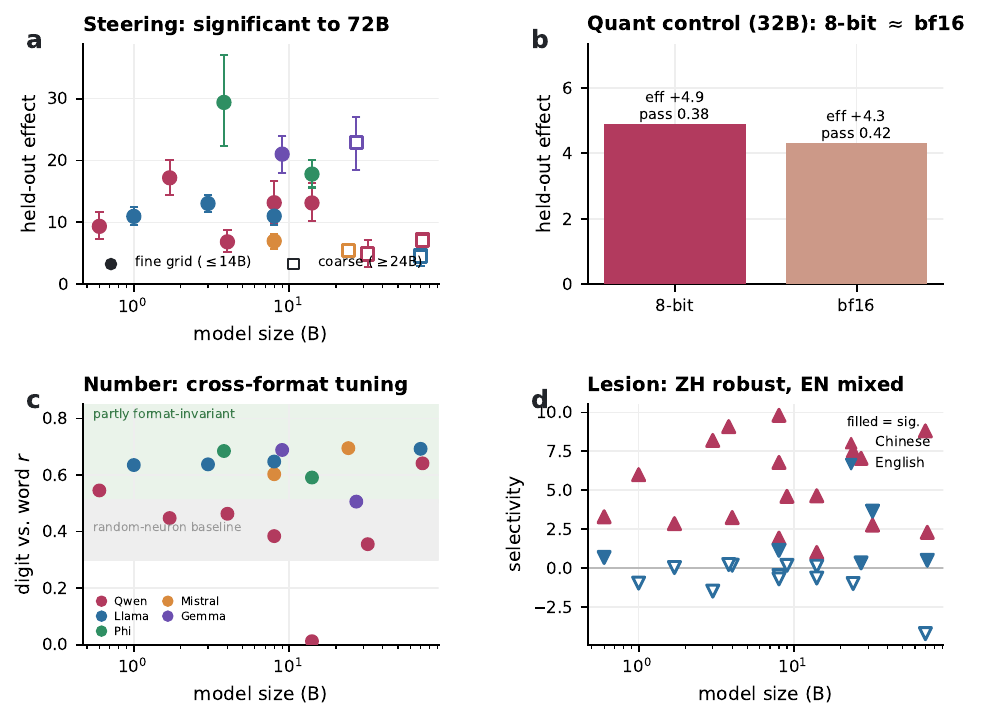}
\caption{\textbf{Cross-family audit at scale ($0.6$--$72$B), from the large-model controls.}
(a)~Held-out steering effect stays significantly positive at every scale (filled $=$ fine grid
$\leq\!14$B; open $=$ coarser grid $\geq\!24$B). (b)~Steering quantization control: Qwen3-32B in
$8$-bit vs.\ bf16 is nearly identical, so precision is not a confound. (c)~Digit-vs-word
cross-format tuning correlation by family (colour legend in panel): the non-Qwen families sit
largely in the ``partly format-invariant'' band ($r>0.5$, an arbitrary cutoff) while Qwen sits
mostly below---their bell units are cross-format consistent, Qwen's are token-like, to $72$B. (d)~Lesion selectivity: Chinese (up) is significant
(filled) in every model; English (down) is significant in only some---the dissociation is
model-dependent.}
\label{fig:scale}
\end{figure*}

The per-model control numbers behind every claim (Table~\ref{tab:results}, in the main text) and the
large-model plot (Fig.~\ref{fig:scale}) are produced by the procedures described below.

\begin{figure}[t]\centering
\includegraphics[width=\columnwidth]{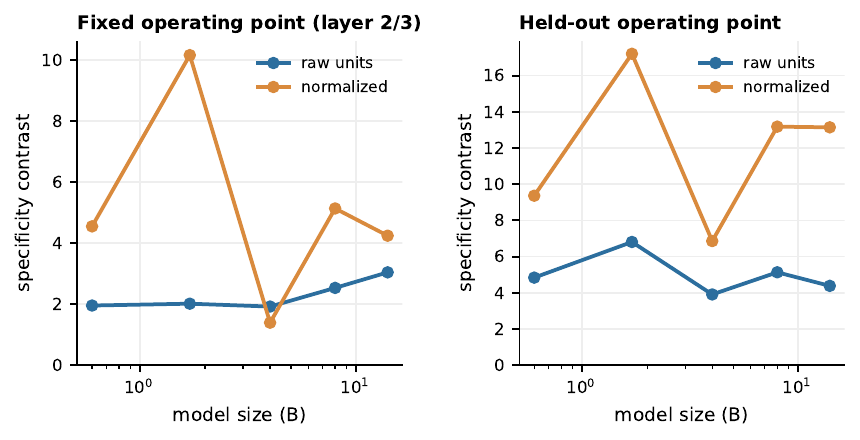}
\caption{\textbf{Factor decomposition: which measurement choice manufactures the apparent scaling?}
Holding the metric (specificity contrast) and concept set ($24$ concepts) constant, we vary two
factors: unit normalization (raw vs.\ $\|h\|$-normalized) and operating-point selection (fixed
$L{\cdot}2/3$ vs.\ held-out). None of the four cells shows a significant scaling trend (all $95\%$ CIs
include $0$). The apparent scaling in the naive pipeline (Fig.~\ref{fig:main}b) was therefore
specific to the compound of raw units, first-token readout, and fixed operating point, not
attributable to any single factor.}
\label{fig:factor}
\end{figure}
\begin{figure}[t]\centering
\includegraphics[width=0.62\columnwidth]{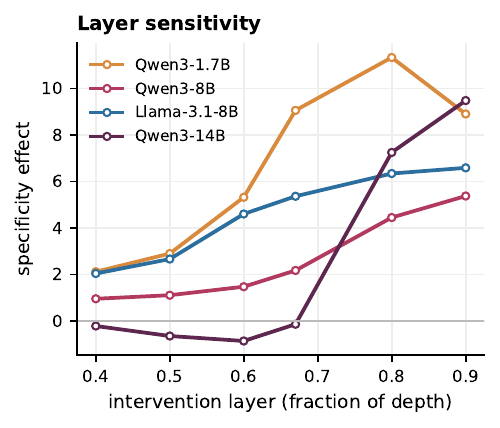}
\caption{\textbf{Layer sensitivity.} Specificity-controlled steering effect (at $c{=}1$) vs.\
the intervention layer, for four models; the effect increases with depth and peaks in late
layers, so a fixed mid-layer choice can misstate it.}
\label{fig:layer}
\end{figure}
\begin{figure}[t]\centering
\includegraphics[width=0.62\columnwidth]{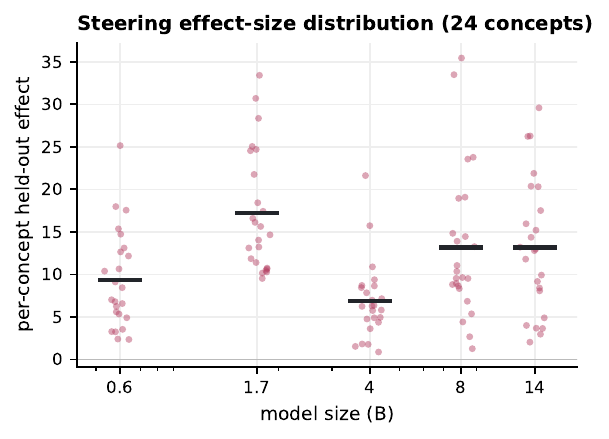}
\caption{\textbf{Steering effect size, not just pass-rate.} Per-concept held-out specificity
effect ($24$ concepts; bar $=$ mean) across the Qwen3 ladder; the distributions overlap across
scale.}
\label{fig:steerdist}
\end{figure}

\textbf{Steering strength.} Fig.~\ref{fig:main}c gives the specificity effect vs.\
$c\in\{0,0.25,0.5,1,2,4,8\}$ (in $\|h\|$ units) per model; the inverted-U peak location varies
by model, underscoring why a single fixed strength is not comparable.
\textbf{Layer.} Fig.~\ref{fig:layer} sweeps the intervention layer over
$\{0.4,0.5,0.6,0.67,0.8,0.9\}\times$depth.
\textbf{Held-out operating-point selection (Fig.~\ref{fig:main}d).} We select the intervention
(layer, strength) exactly as exp02 selects its probe layer and $\alpha$: by held-out
performance. For each Qwen3 model we compute the per-concept specificity effect at every cell of
a layer$\times$strength grid (layer $\in\{0.4,0.6,0.8\}\times$depth, strength
$c\in\{0.25,0.5,1,2,4\}$) over \emph{$24$ concepts in $5$ categories}, then $4$-fold over the
concepts: pick the cell with the largest mean effect on the select folds, score the disjoint
held-out fold. Held-out pass-rate is $\{0.71,1.00,0.79,0.79,0.75\}$ across $0.6$--$14$B, each
$95\%$ bootstrap CI (over concepts) above chance; the continuous effect is significantly
positive throughout; the slope of effect against $\log_2$ size is $+0.31$, $95\%$ CI
$[-0.11,+0.73]$: no \emph{significant} trend, but the CI does not exclude a moderate positive
slope (an underpowered null; excluding a $+0.3$/doubling trend at this effect variance would need
roughly $3$--$4\times$ more concepts or models). Every model selects layer $\approx\!0.8$. Category
means (Qwen3-8B): object $+20.4$, person $+17.4$, city $+14.6$, animal $+8.2$, abstract $+7.5$.
By contrast the hand-picked $(2/3,1.0)$ cell gives an erratic $\{0.50,0.88,0.25,0.50,0.25\}$---
the fixed-layer ``scaling'' is an artifact of an arbitrary operating point. \emph{Layer vs.\
strength drift.} Decomposing the two sources: the optimal layer is consistently $\approx\!0.8$
across all sizes (Fig.~\ref{fig:layer}), so the hand-picked $2/3$ is a \emph{uniform} layer mismatch
(too shallow for every model), not a scale-specific drift; the optimal \emph{strength} varies
more by model (the dose--response peaks shift, Fig.~\ref{fig:main}c). The erratic fixed-point is thus
driven mainly by the layer mismatch with a smaller strength component---but either way, a single
fixed $(\text{layer},\text{strength})$ is not comparable across models.
Fig.~\ref{fig:steerdist} shows the full per-concept effect-size distribution behind these pass-rates
(not just the binary metric): the distributions overlap heavily across scale, consistent with
the null trend.
\textbf{Large models and quantization.} On a coarser grid (layer $\in\{0.67,0.8\}$, strength
$\in\{0.5,1,2\}$) the held-out effect stays significantly positive at $24$--$72$B, with a
bootstrap-over-concepts $95\%$ CI strictly above $0$ for every model: Gemma-2-27B $+22.9\,[+18.5,+27.1]$, Mistral-24B
$+5.5\,[+4.5,+6.4]$, Qwen3-32B $+4.9\,[+2.7,+7.2]$, Qwen2.5-72B $+7.1\,[+6.0,+8.3]$, Llama-3.1-70B
$+4.5\,[+3.0,+6.1]$. We rely on this effect CI, not the pass-rate: the coarse grid gives lower
absolute pass-rates (e.g.\ $0.38$ at $32$B) that are \emph{not} comparable to the fine $14$B scan,
so pass-rate is not used as significance evidence at $\ge\!24$B. A steering \emph{quantization
control}---Qwen3-32B in $8$-bit vs.\ bf16---is nearly identical (effect $+4.9$ vs.\ $+4.3$;
overlapping CIs), so precision does not drive the steering result. Because the grid is coarse and
mixes families, we read significance and precision-invariance here, not a fine trend.
\textbf{Lesion fraction.} Fig.~\ref{fig:lesion}b sweeps the ablated fraction over
$\{0.05,0.1,0.2,0.5,1\}\%$ of neurons; the Chinese effect is stable across it.
\textbf{Probe regularization and seeds.} Map probes use CV-selected ridge $\alpha$; nulls use
$200$ label shuffles.
\textbf{Magnitude shape (shape-agnostic selection).} To remove the bias of selecting neurons by
\emph{linear} correlation with number---which can only return monotonic units---we select the
top $30$ neurons by held-out $R^2$ of a \emph{quadratic} fit (even/odd number split), then class
a neuron as bell if its full-data quadratic beats the linear fit by $>0.15$ $R^2$ with an
interior extremum ($5\!\le\!$ vertex $\!\le\!36$). The bell fraction is then $0.00$--$0.43$
(rare in Qwen, $13$--$43\%$ in Llama/Phi/Mistral) versus $0.00$ everywhere under linear
selection; well-tuned bell-shaped number neurons are present in \emph{all} models.
\textbf{On thresholds.} Our four-cell verdict introduces several cutoffs (quadratic-gain $>0.15$,
interior vertex, cross-format $r>0.5$, pass-rate), so we note which conclusions depend on them. Two
do not: the steering \emph{units} artifact follows from residual-norm growth regardless of any
cutoff, and the world map clears its shuffle null by a wide margin at every regularizer we tried. The
central magnitude claim---that bell-shaped units \emph{exist} and are hidden by linear selection---is
also threshold-insensitive in direction, since linear selection returns $0\%$ bell \emph{by
construction} for any positive quadratic-gain cutoff; only the precise bell \emph{fraction} moves
with the cutoff, which is why we report the magnitude verdict as \emph{partial}/unresolved rather
than as a definite shape. The lesion verdict is a null/instability result that only weakens under
looser thresholds.
\textbf{Magnitude cross-format test.} To separate genuine numerosity from digit-token identity,
we present each number as a digit (``$17$'') and a spelled-out word (``seventeen''), and for the
bell neurons selected in the digit format correlate their digit-format and word-format tuning
curves. Mean cross-format $r$ is $0.59$--$0.70$ for Llama/Phi/Mistral ($73$--$83\%$ of bell units
$>0.5$), above each model's random-neuron baseline ($0.30$--$0.51$), but $0.01$--$0.64$ for Qwen (most below $0.55$; Qwen2.5-72B is the exception at $0.64$): the
bell units are \emph{partly} format-invariant in the former (consistent across digit vs.\ word),
weak or token-specific in the latter.
\textbf{Lesion bootstrap.} With $44$ sentences/language (train $32$, test $12$) and $3000$
test-set bootstrap resamples, Chinese selectivity has a $95\%$ CI strictly $>0$ in all $17$ tested
models; English selectivity is not significant (CI includes $0$) in $12$ and significantly positive
in only $5$ (Qwen3-0.6B, Ministral-8B, Qwen3-32B, Qwen2.5-72B, Gemma-2-27B). A full double dissociation (both
halves significant, positive) holds in $5/17$.
\textbf{Lesion attribution robustness.} Re-selecting language neurons by \emph{activation
magnitude} (mean max-token activation difference between languages) rather than
gradient$\times$activation flips the asymmetry: over the $9$ models tested both ways, Chinese is then
significant in $7/9$ (two Qwen models fail) and English in $7/9$. The specific ``Chinese-robust, English-null''
pattern is thus attribution-dependent; only the existence of \emph{some} localizable
language structure is robust to the method change.

\section{Qualitative examples}\label{app:qual}
\begin{figure*}[t]\centering
\includegraphics[width=0.86\textwidth]{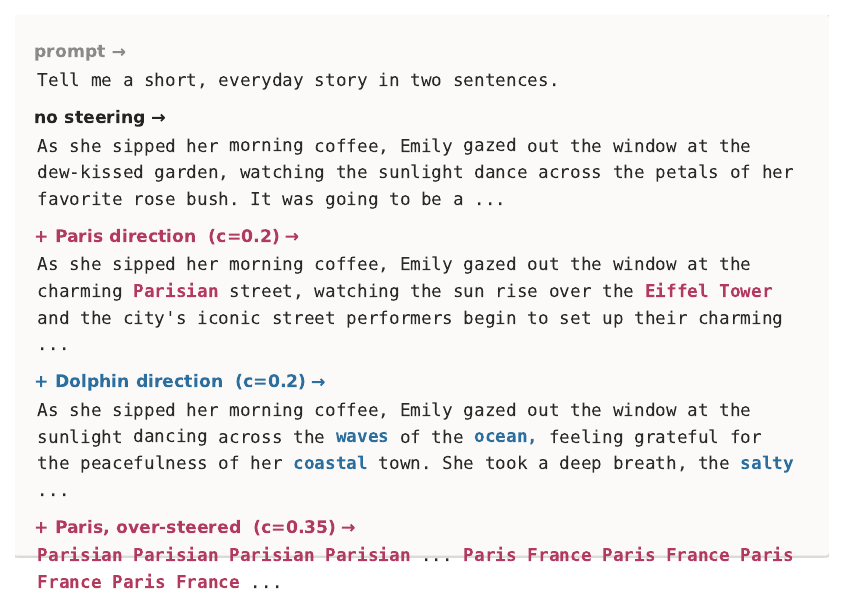}
\caption{\textbf{What steering does, in words} (real Llama-3.1-8B generations, one fixed prompt). A
concept direction added to the residual stream injects that concept's vocabulary (highlighted) while
the sentence stays coherent at a moderate coefficient. The last row shows over-steering
($c\!=\!0.35$): the same direction now dominates and the text collapses into repetition, the
free-generation face of the inverted-U dose-response in Fig.~\ref{fig:main}c. The fluent band is
concept-dependent, so a single fixed coefficient is not comparable across concepts.}
\label{fig:steerexample}
\end{figure*}

This appendix gives one concrete, real example per experiment, so the aggregate statistics have a
face. Fig.~\ref{fig:steerexample} shows what a steering intervention actually does to the text. Every row is a
real, greedy-decoded output of Llama-3.1-8B for the \emph{same} prompt, with a concept direction
added to the residual stream at layer $\approx\!0.8$ depth (Methods); the injected concept
vocabulary is highlighted. Unsteered, the model writes a neutral vignette. Adding the \emph{Paris}
direction at a moderate strength ($c\!=\!0.2$) bends the same story toward Paris (``Parisian
street,'' the ``Eiffel Tower'') while staying fluent, and the \emph{Dolphin} direction pulls the
same opening to the sea (``waves of the ocean,'' ``salty air'').

\begin{figure}[t]\centering
\includegraphics[width=\columnwidth]{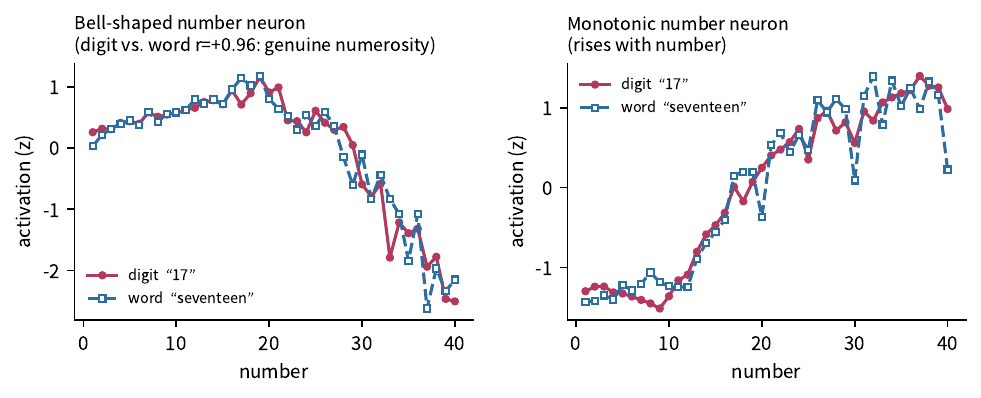}
\caption{\textbf{Real number-neuron tuning curves} (Llama-3.1-8B). A bell-shaped and a monotonic
unit; each neuron's digit and word tuning curves nearly coincide ($r=0.96$), indicating magnitude
tuning beyond token identity.}
\label{fig:numberexample}
\end{figure}

\paragraph{Number neurons keep their tuning across formats.} Fig.~\ref{fig:numberexample} plots two real
number-tuned neurons in Llama-3.1-8B, one \emph{bell-shaped} (peaking near $18$) and one
\emph{monotonic}. Overlaying the curve elicited by digits (``$17$'') and by spelled-out words
(``seventeen'') shows the two nearly coincide (cross-format $r=0.96$), evidence that the unit encodes
numerical magnitude rather than a specific digit token.

\begin{figure}[t]\centering
\includegraphics[width=\columnwidth]{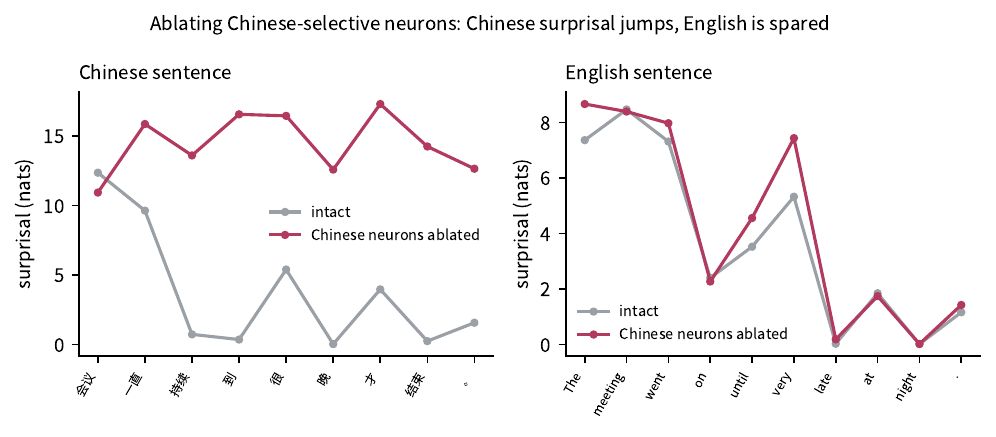}
\caption{\textbf{A language lesion, token by token} (Llama-3.1-8B). After ablating
Chinese-selective neurons, surprisal on a Chinese sentence rises sharply while an English sentence is
spared.}
\label{fig:lesionexample}
\end{figure}

\paragraph{Lesioning Chinese neurons is language-specific.} Fig.~\ref{fig:lesionexample} shows per-token
surprisal on a held-out Chinese and English sentence before and after ablating the top
Chinese-selective neurons. Chinese surprisal jumps on essentially every token (mean $+10.7$ nats),
whereas the English sentence is almost unchanged ($+0.5$ nats): a single clean example of the
selective loss the aggregate statistics summarize.

\begin{figure}[t]\centering
\includegraphics[width=\columnwidth]{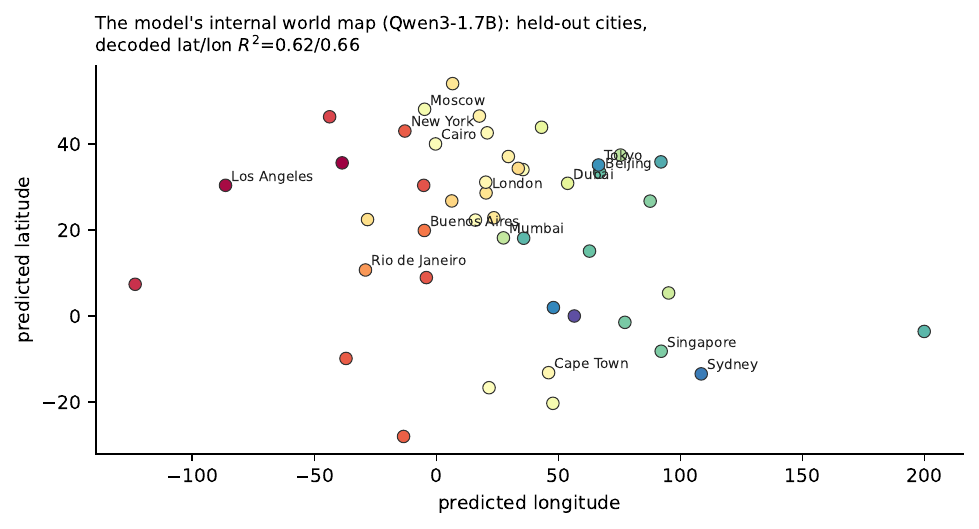}
\caption{\textbf{The recovered world map} (Qwen3-1.7B, a representative small model). Cross-validated
predicted coordinates of held-out cities; colour encodes true longitude. Even at $1.7$B, geography
is recovered ($R^2=0.62/0.66$ for latitude and longitude).}
\label{fig:mapexample}
\end{figure}

\paragraph{The world map is legible.} Fig.~\ref{fig:mapexample} plots the cross-validated latitude and
longitude a ridge probe recovers for each held-out city from the residual-stream vector of Qwen3-1.7B, a
representative small model. Even at $1.7$B the recovered coordinates reproduce global geography (Los
Angeles west, Tokyo and Beijing east, Moscow north, Cape Town south), the one parallel that survives
every control.

\paragraph{Representative stimuli.} Concept directions are built from short sentences, for
\emph{Paris} ``Paris is the capital of France'' and ``We flew into Paris in spring,'' scored on the
target words \{Paris, France, Eiffel, French\} against a same-category control (\emph{Tokyo}).
Number tuning uses prompts ``The number is $n$'' for $n=1,\dots,40$ in both digit (``$17$'') and word
(``seventeen'') form. Language lesioning uses $44$ matched sentence pairs per language, for example
the English ``The weather is lovely today, let's take a walk in the park'' and its Chinese
translation of the same meaning.

\section{Reproducibility}\label{app:details}
All models are open-weight (Hugging Face); interventions use forward hooks on the residual
stream and MLP layers. Code, stimuli, exact model revisions, and the JSON results behind every
figure are released (see footnote~1). We evaluate $17$ open checkpoints, $0.6$--$72$B: Qwen3
$\{0.6,1.7,4,8,14,32\}$B and Qwen2.5-$72$B; Llama-3.2 $\{1,3\}$B, Llama-3.1-$8$B, and Llama-3.1-$70$B;
Phi-3.5-mini and Phi-4; Ministral-$8$B and Mistral-Small-$24$B; and Gemma-2 $\{9,27\}$B. Checkpoints $\le\!14$B run in
bf16 on a single GPU; $24$--$72$B run quantized ($8$-bit where they fit an $80$\,GB card, else
$4$-bit for $70$--$72$B), with a $32$B bf16-vs-$8$-bit control confirming the coding results are
precision-independent. \textbf{Memory-efficient lesion.} The gradient$\times$activation
attribution originally exceeded memory on $\ge\!14$B models because the backward pass allocated
a full weight-gradient buffer; freezing all weights and marking only the recorded activations
\texttt{requires\_grad} removes that buffer ($\sim$model-size memory saved) while leaving the
attribution \emph{numerically identical}, so all $17$ models are now covered.
\end{document}